\documentclass[times,twocolumn,final,authoryear]{elsarticle}

\usepackage{ycviu}
\usepackage{framed,multirow}

\usepackage{amssymb}
\usepackage{latexsym}

\usepackage{url}
\usepackage{xcolor}
\definecolor{newcolor}{rgb}{.8,.349,.1}

\usepackage{graphicx}
\usepackage{amsmath}
\usepackage{amssymb}
\usepackage{booktabs}
\usepackage[noend]{algpseudocode}
\usepackage{makecell}

\usepackage{enumitem}
\usepackage{comment}
\usepackage{multirow}

\DeclareMathOperator*{\argmin}{arg\,min}

\makeatletter
\DeclareRobustCommand\onedot{\futurelet\@let@token\@onedot}
\def\@onedot{\ifx\@let@token.\else.\null\fi\xspace}

\def\eg{\emph{e.g}\onedot} 
\def\ie{\emph{i.e}\onedot}

\usepackage{accents}

\usepackage{xcolor}
\usepackage{multirow}

\makeatother

\usepackage{xspace}
\makeatletter
\DeclareRobustCommand\onedot{\futurelet\@let@token\@onedot}
\def\@onedot{\ifx\@let@token.\else.\null\fi\xspace}

\def\eg{\emph{e.g}\onedot} 
\def\ie{\emph{i.e}\onedot}

\makeatother

\makeatletter
\newcommand{\removelatexerror}{\let\@latex@error\@gobble}
\makeatother

\usepackage{epsfig}
\usepackage{overpic}
\usepackage{float}
\floatstyle{plaintop}
\restylefloat{table}
\usepackage[export]{adjustbox}
\usepackage{multirow}
\usepackage{subfig}
\usepackage{amsmath}

\usepackage[linesnumbered,ruled,vlined]{algorithm2e}
\SetKwInput{KwInput}{Input}                
\SetKwInput{KwOutput}{Output}              

\usepackage{hyperref}

\newcommand*{\rev}{\textcolor{black}}
\let\oldnl\nl
\newcommand{\nonl}{\renewcommand{\nl}{\let\nl\oldnl}}

\journal{Computer Vision and Image Understanding}

\begin{document}

\begin{frontmatter}

\title{Self-Supervision \& Meta-Learning for One-Shot Unsupervised Cross-Domain Detection}

\author[1,2]{Francesco \snm{Cappio Borlino}\corref{cor1}} 
\cortext[cor1]{Corresponding author: 
  }
\ead{francesco.cappio@polito.it}
\author[1]{Salvatore \snm{Polizzotto}}
\author[1]{Barbara \snm{Caputo}}
\author[1,2]{Tatiana \snm{Tommasi}}

\address[1]{Politecnico di Torino, Corso Duca degli Abruzzi 24, 10129 Torino, Italy}
\address[2]{Italian Institute of Technology, Italy}

\begin{abstract}
Deep detection approaches are powerful in controlled conditions, but appear brittle and fail when source models are used off-the-shelf on unseen domains. Most of the existing works on domain adaptation simplify the setting and access jointly both a large source dataset and a sizable amount of target samples. However this scenario is unrealistic in many practical cases as when monitoring image feeds from social media: only a pretrained source model is available and every target image uploaded by the users belongs to a different domain not foreseen during training. We address this challenging setting by presenting an object detection algorithm able to exploit a pre-trained source model and perform unsupervised adaptation by using only one target sample seen at test time. Our multi-task architecture includes a self-supervised branch that we exploit to meta-train the whole model with single-sample cross-domain episodes, and prepare to the test condition. At deployment time the self-supervised task is iteratively solved on any incoming sample to one-shot adapt on it.
We introduce a new dataset of social media image feeds and present a thorough benchmark with the most recent cross-domain detection methods showing the advantages of our approach.
\end{abstract}


\end{frontmatter}

\section{Introduction}
\label{sec:introduction}
\begin{figure}[tb]
    \centering
    \includegraphics[width=0.48\textwidth]{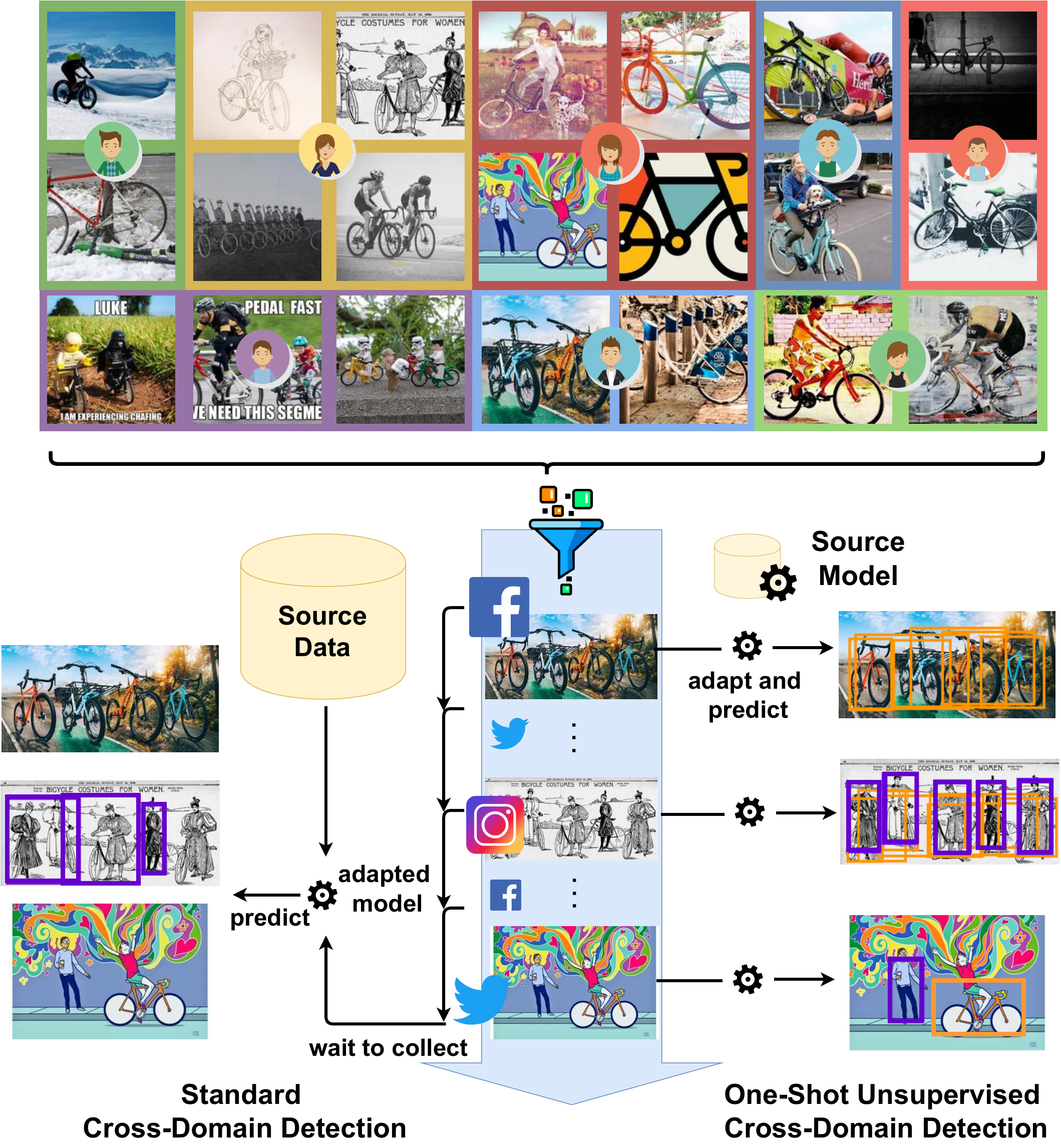}
    \caption{\rev{\textbf{Top}: examples of our newly collected Social Bikes dataset. Every user has its own preferred image style. 
    The images appear shuffled on social media feeds. \textbf{Bottom}: two different strategies to tackle the social media monitoring task. \textbf{Bottom Left}: standard cross-domain detection. Existing algorithms need to access the source data and wait to collect a batch of target samples. \textbf{Bottom Right}: one-shot unsupervised cross-domain detection. OSHOT and FULL-OSHOT exploit the source model without accessing the source data. The detection model is adapted on a single target image and predicts on it.}}
    \label{fig:newteaser} \vspace{-4mm}
\end{figure}

Despite impressive progress in object detection over the last years, reliably localizing and recognizing objects across visual domains is still an open problem. Indeed, most of the existing detectors rely on deep features learned from large amount of labeled training data usually drawn from a specific \emph{source} distribution and suffer from severe performance degradation when applied on images sampled from a different \emph{target} domain. This hinders the deployment of detection models in real-world conditions.
Consider for example the task of social media monitoring (see Figure \ref{fig:newteaser}): the images are posted on multiple platforms
by a large variety of users, each with his/her own personal taste in the choice of style and post-processing filters which may change in time. Even when images contain instances of the same object category, they are acquired in different contexts, under different viewpoints and illumination conditions. In other words, \emph{each image comes from a different visual domain, distinct from the visual domain where the detector has been trained}. 
\rev{
This scenario poses several key 
challenges: (1) the model faces a stream of test samples, which 
come from different target domains. Thus, there is no guarantee that the images received from time $t+1$ will be drawn from the same distribution as those observed up to time $t$; (2) collecting a batch of test samples for adaptation would cause an unacceptable prediction delay.
Moreover, it is pointless: the target domain will suddenly change deprecating the model; (3) the annotated source images on which the detector is trained are not accessible at deployment time (as they might be proprietary), hence the adaptation should happen in a source-free fashion. The best way to tackle all these challenges is to adapt the source model on each test sample just before performing the prediction. 
}

Our work addresses this setting that we named \emph{One-Shot Unsupervised Cross-Domain Detection} in \citep{oshot_eccv20}. 
\rev{The method OSHOT proposed in the same work presents a clear asymmetry between the training and testing procedures. When learning the source model, detection and self-supervision run jointly and support each other to obtain a robust representation. However, for each test sample, self-supervised learning is first applied alone for some fine-tuning adaptation steps on a region of interest roughly located via pseudo-labeling, and it is only then followed by detection. We propose to remove this asymmetry, which means designing a source training strategy that prepares the model to the specific condition that it will face during deployment. 
Meta-learning perfectly serves this  purpose, making the source model ready to be transferred on one single sample at inference time.}

\rev{Overall, the contributions of this work can be summarized as follows:
(1) we introduce FULL-OSHOT that leverages a novel meta-learning formulation to better combine the main supervised detection task with the self-supervised auxiliary objective. As well as OSHOT, the new variant does not need access to the source data for adaptation, and it is better suited for the one-shot unsupervised cross-domain scenario because it mimics the inference conditions at training time. (2) We present an extended version of the Social Bikes dataset, created as testbed for the challenging task of object detection on social media feeds. We passed from 30 to 530 samples collected from Twitter, Instagram and Facebook by searching the \#bike tag. The domain variability covered by this new dataset exceeds that of data collections previously used for standard cross-domain detection tasks. 
(3) We run a thorough experimental benchmark by comparing FULL-OSHOT with the most recent adaptive detection algorithms
\citep{Saito_2019_CVPR,diversifymatch_Kim_2019_CVPR,xuCVPR2020,wu2021vector} and one-shot style-transfer based unsupervised learning technique \citep{Cohen_2019_ICCV},
achieving the new state-of-the-art. 
(4) We go beyond showing qualitative detection outputs by presenting a detailed study on the prediction errors via the toolbox proposed in \citep{tide-eccv2020}.  Moreover, we carefully assess the role of the inner components of our approach through several ablation experiments. 
}

\rev{Code, implementation details and qualitative results of our FULL-OSHOT are available at \url{https://github.com/FrancescoCappio/OSHOT-meta-learning}.} 
\section{Related Work}
\label{sec:related}
\noindent\textbf{Cross-Domain Object Detection.} 
Existing deep learning based object detectors perform remarkably well \citep{ren2015faster,Lin_2017_ICCV}, however their robustness across visual domains remains a major issue. 
Unsupervised domain adaptation methods attempt to close the domain gap between the annotated source on which learning is performed, and the target samples on which the model is deployed. \rev{\emph{The standard adaptive setting assumes the availability of all the unlabeled target test data at training time.}
Cross-domain analysis has been mainly studied in the object recognition context \citep{DA_theory,AFN,jigsaw_pami}, with a certain number of strategies developed also for object detection.}
One adaptive strategy consists in including \emph{feature alignment} modules at different internal stages of the deep architecture \citep{Chen_2018_CVPR}.
The Strong-Weak method  (SW, \cite{Saito_2019_CVPR}) proposed a balanced alignment with strong global and weak local feature adaptation. 
The SW-ICR-CCR method \citep{xuCVPR2020} includes an image-level multi-label classifier and a module imposing consistency between the image-level and instance-level predictions. The recent approach ICCR-VDD \citep{wu2021vector} exploits vector decomposition to separate domain-invariant and domain-specific representation with the former used to extract object proposals.
Another group of works developed \emph{pixel-level adaptation} methods which modify source images to resemble those of the target. The Domain-Transfer approach (DT, \cite{inoue2018cross}) was the first to apply this strategy for object detection. 
More recently Div-Match  \citep{diversifymatch_Kim_2019_CVPR} re-elaborated the idea of domain randomization \citep{Tobin2017DomainRF} to produce three extra source variants with which the target can be aligned through an adversarial multi-domain discriminator.
Finally, pseudo-labeling, also known as \emph{self-training}, uses the output of the source detector as a coarse annotation for the unlabeled target samples which are then included in the supervised training \citep{kim2019selftraining,robust_Khodabandeh_2019_ICCV}. 
\rev{Other cross domain detection settings have received scarce attention, with a single work studying \emph{domain generalization} \citep{Wang_2019_CVPR}. There the target data are not available at training time, but it is essential to leverage \emph{multiple source domains}.}

\noindent\textbf{Adaptive Learning on a Budget.}
When dealing with domain shift, learning on a target budget becomes extremely challenging. Only few attempts have been done to reduce the target cardinality \rev{and they all focus on object classification}. \cite{fewshotNIPS17} considered \emph{few-shot supervised domain adaptation} where the few target samples available are fully labeled. \rev{\cite{Liu_2021_ICCV} proposed the \emph{multi-domain supervised few-shot classification task}, where the goal is learning new categories from few labeled samples that can be drawn from an unseen domain. 
}
\cite{Cohen_2019_ICCV} \rev{addressed} \emph{one-shot unsupervised style transfer} with a large source dataset and a single unsupervised target image. They developed BiOST, a time-costly autoencoder-based method whose goal is image generation with no discriminative purpose.
\rev{The work of \cite{efrostesttimeICML2020} was concurrent to our \citep{oshot_eccv20} and proposed \emph{test-time training}, which is analogous to our one-shot unsupervised learning setting, but for object classification.}
A related scenario is that of \emph{online domain adaptation} where unsupervised target samples belong to a single coherent domain and are initially scarce, but accumulate in time \citep{Hoffman_CVPR2014,mancini2018kitting}. 

\noindent\textbf{Self-Supervised Learning.} Unlabeled data is rich in structural information and self-supervised learning aims at capturing it, to then serve as a pre-training step for different downstream tasks.
Recently \cite{asano20a-critical} have shown the potential of a self-supervised model learned from a single image. 
Several works have also indicated that self-supervision supports generalization when combined with supervised learning in a multi-task framework  \citep{jigsaw_pami, Xu2019SelfsupervisedDA,AAlliegro_ICPR2021,Bucci2019TacklingPD}. 
\rev{Indeed, the joint-learning procedure extracts multiple and complementary information from the data, which are integrated in an enhanced feature embedding, in line with the Multiple Knowledge framework discussed in  \citep{yang_multiple_2021}.}

\begin{figure*}[tb]
    \centering
    \includegraphics[width=0.85\textwidth]{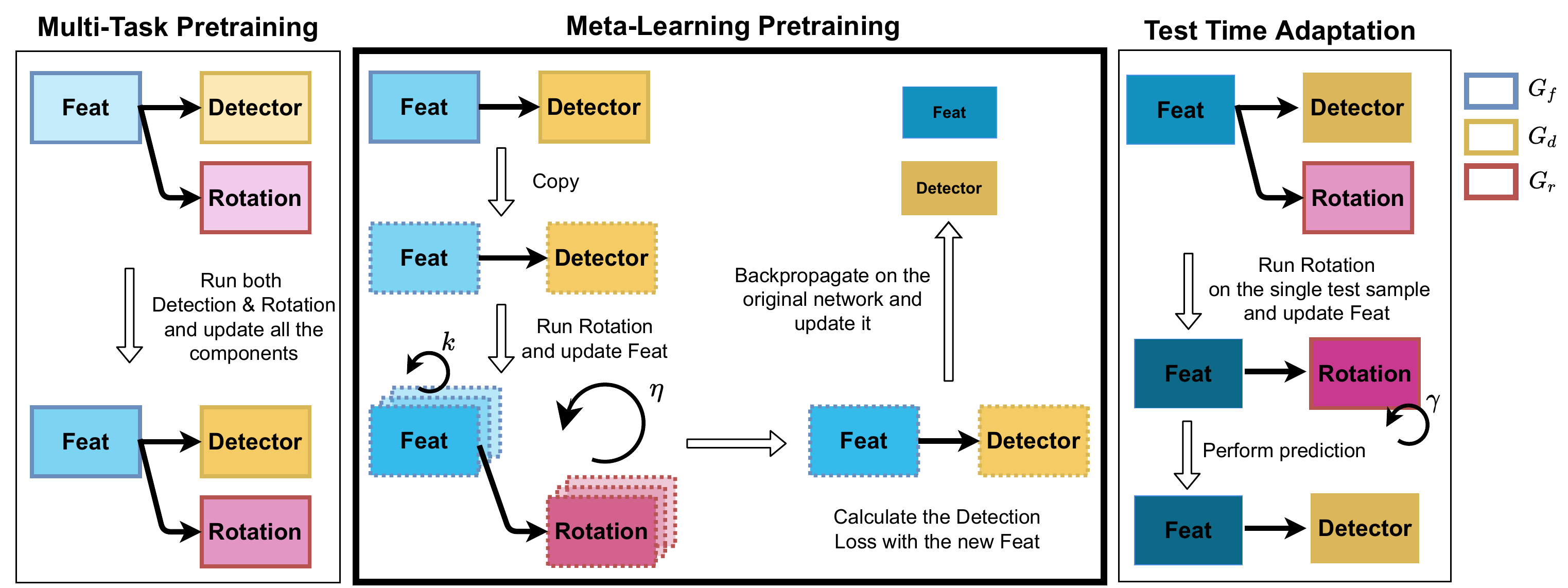}
    \caption{Visualization of the complete FULL-OSHOT proposed approach. The first phase, inherited from OSHOT, is a multi-task pretraining in which all the modules are updated through the rotation and detection losses. The second pretraining stage, introduced for FULL-OSHOT, exploits a meta-learning with the rotation recognition as the inner optimization task to prepare the network for the test-time adaptation stage. In this last phase both the rotation and feature extractor modules are updated by performing the self-supervised task iteratively on a single test sample. The adapted feature extractor is finally used to predict on the same test image. Each change in color shades indicates an update of a module. 
    We use dotted lines to highlight the components optimized in the meta-learning loop.}
   \label{fig:ourapproach}\vspace{-3mm}
\end{figure*}

\noindent\textbf{Meta-Learning.} 
The objective of meta-learning is to enable a model for fast adaptability. 
A well known strategy is that proposed by \cite{maml-finn17a}: the \emph{inner} learning loop solves a standard supervised task, while the \emph{outer} meta-learning loop updates the base model by observing multiple episodes of the standard task to accomplish a higher level objective as generalization or increasing learning speed. This technique allows adaptation to novel tasks with scarce supporting examples and has been largely used for few-shot learning  \citep{NIPS2017_fewshotproto,NIPS2016_matching,rusu2018metalearning}. Various kind of meta-knowledge as losses \citep{featurecritic}, regularization functions \citep{metareg2018} and data augmentation \citep{Tseng2020Cross-Domain} can be (meta) learned to maximize the model robustness by using a validation domain different from the training one.

\smallskip \rev{
\noindent\textbf{In this work we propose to use meta-learning to improve how the supervised and self-supervised tasks are combined at training time for one shot unsupervised cross-domain detection.}
For each training sample we use its augmented views to simulate cross-domain learning episodes for the self-supervised task. This operates as inner optimization loop and defines the feature representation. Globally, 
the network parameters are updated (outer loop) to produce the best detection performance by building on the adapted features.
This training strategy mimics what will happen at test time and provides a model better able to adapt on a single sample.}

\section{Method}
\label{sec:method}
\noindent\rev{\textbf{Preliminaries and problem specification.} The training dataset is composed of $N$ annotated samples of the source domain $S=\{x_{i},y_{i}\}_{i=1}^N$, and can be used for model training only, \ie it will not be available during deployment. At inference time, test samples arrive one by one: we want to perform detection on a single image $x^t$, with $t$ being any target domain not available at training time.} Here the structured labels $y=(c,b)$ describe class identity $c$ and bounding box location $b$ in each image $x$. 

In \citep{oshot_eccv20} we \rev{introduced}  
OSHOT: a deep multi-task method that included a \emph{pretraining} and a \rev{test-time \emph{adaptation}} phase. The former involves the source \rev{and consists in} optimizing jointly the detection and \rev{the} 
auxiliary self-supervised rotation recognition objective. In the latter stage, the network features are updated on the single target sample by focusing only on the rotation task before performing a prediction. 
\rev{Moreover, the approach exploits \emph{self-training} in a \emph{cross-task} fashion: a rough object bounding box is provided to the auxiliary rotation task that will focus on that image area.}
The basic detector is Faster R-CNN \citep{ren2015faster}, that has three main components: an initial block of convolutional layers, a region proposal network (RPN) and a region-of-interest (ROI) based classifier. The bottom layers transform any input image $x$ into its convolutional feature map $G_{f}(x|\theta_{f})$ where $\theta_{f}$ parametrizes the feature extraction model. The feature map is then used by RPN to generate candidate object proposals. Finally the ROI-wise classifier $G_d(\cdot|\theta_d)$ predicts the category label and object position from the feature vectors obtained via ROI-pooling.
\rev{We propose here to extend the original OSHOT through a tailored 
meta-learning pretraining phase, designed to prepare the model to be effectively transferred on one single sample, and to produce good results after a few adaptation iterations on it.
}

\noindent\textbf{Multi-task Pretraining} (Figure \ref{fig:ourapproach}, left).
The detection head and the feature extractor are trained jointly \rev{by minimizing the} 
loss function $\mathcal{L}_{d}(G_{d}(G_{f}(x|\theta_{f})|\theta_{d}), y)$ which evaluates cost errors for both object classification and regression of the identified bounding boxes.
For the auxiliary rotation task, each training image $x$ is transformed via a rotation operator $R(x, \alpha)$, where $\alpha= q \times 90^{\circ}$ \rev{is} the orientation with  $q \in \{0,\ldots,3\}$. We indicate the obtained set of samples as $\{R(x)_j, q_j\}_{j=1}^M$, where we dropped the $\alpha$ for simplicity. 
We \rev{refer to} the auxiliary rotation classifier and its parameters respectively as $G_{r}$ and $\theta_{r}$. 
The overall objective of the multi-task  
model is:\vspace{-1mm}
\begin{equation}
    \argmin_{\theta_{f}, \theta_{d}, \theta_{r}} 
    \mathcal{L}_{d}(G_{d}(G_{f}(x|\theta_{f})|\theta_{d}),y) \ + 
    \lambda \mathcal{L}_{r}(G_{r}(G_{f}(R(x)|\theta_{f})|\theta_{r}), q)~,
\end{equation}
where $\mathcal{L}_{r}$ is the cross-entropy loss \rev{and $\lambda$ is used to control the relative importance of the auxiliary tasks.} 
The shared feature map $G_{f}(x|\theta_{f})$ is thus learned under the synchronous guidance of both the detection and rotation objectives. 
\rev{During training, $G_{r}$ exploits} 
the ground truth location of each object and 
\rev{selects} 
features from its bounding box in the original map $G_{r}(\cdot |\theta_r) = \mbox{FC}_{\theta_r}(boxcrop(\cdot))$. The $boxcrop$ operation includes pooling to rescale the feature dimension before entering the final FC layer. 
In this way the network focuses 
on the object orientation without introducing noisy information from the background. We randomly pick one rotation angle per instance. 

\noindent\textbf{Meta-Learning Pretraining} (Figure \ref{fig:ourapproach}, center). Multi-task learning is appealing for deep learning regularization and including a self-supervised task has the advantage of waiving any extra data annotation cost. Still, our main interest remains on detection, while rotation recognition should be considered as a secondary task. To manage this role for rotation, and to better fit to the unlabeled one-shot scenario on a new domain faced at test time, we re-formulate the OSHOT model inspired by meta-learning and building over the bi-level optimization process of MAML \citep{maml-finn17a}. Specifically we propose to meta-train the detection model with the rotation task as its inner base learner. The optimization objective can be written as
\vspace{-1mm}
\begin{equation}
\begin{aligned}
    & \argmin_{\theta_{f}, \theta_{d}} \frac{1}{K}\sum_{k=1}^K \mathcal{L}_{d}(G_{d}(G_f(x_k|\theta'_f)|\theta_{d}),y)  \\
    \text{s.t.} \  (\theta'_f, \theta'_r)  = \ & \argmin_{\theta_{f}, \theta_{r}} \mathcal{L}_{r}(G_{r}(G_{f}(R(x_k)|\theta_{f})|\theta_{r}), q)~.
\end{aligned}
\label{eq:meta}
\end{equation}
In words, we start by focusing on the rotation recognition task for each source sample $x$ after augmenting it in $k=1,\ldots,K$ different ways. We consider semantic-preserving augmentations (\eg gray-scale, color jittering) and perform multiple learning iterations ($\eta$ gradient-based update steps). This optimization, whose learning objective is reported in the second row of Equation  (\ref{eq:meta}), leads to the update of the feature extractor and rotation classification modules (parameters $\theta'_f$ and $\theta'_r$). The outer meta-learning loop, whose learning objective is in first row of Equation  (\ref{eq:meta}), leverages on it to optimize the detection model over all the $K$ data variants and prepares for generalization and fine-tuning on a single sample. 
To simulate the deployment setting we neglect the ground truth object location for the inner rotation objective and in $G_{r}$ we substitute the $boxcrop$ with $pseudoboxcrop$ obtained through the cross-task self-training procedure detailed in the following paragraph. 

\noindent\textbf{Cross-task self-training.}
Instead of following the self-training standard practice which consists in using the pseudo-labels produced by the source model on the target to update the detector, we exploit them for the self-supervised rotation classifier. With this cross-task self-training we keep the advantage of the self-training initialization, while largely reducing the risks of error propagation due to wrong class pseudo-labels.
We start from the $(\theta_{f},\theta_{d})$ model parameters of the pretraining stage and we get the feature maps from all the rotated versions of the sample $x$,  $G_{f}(\{R(x),q\}|\theta_{f})$, $q={0,\ldots,3}$. 
Only the feature map produced by the original image (\ie $q=0$) is provided as input to the RPN and ROI network components to get the predicted detection $\tilde{y}=(\tilde{c},\tilde{b})=G_{d}(G_{f}(x|\theta_{f})|\theta_{d})$. This pseudo-label is composed by the class label $\tilde{c}$ and the bounding box location $\tilde{b}$. We discard the first and consider only the second to localize the region containing an object in all the four feature maps, also recalibrating the position to compensate for the orientation of each map. 
The $pseudoboxcrop$ operation is used both in the meta-learning phase and in the adaptation one: it guides rotation recognition to focus on object regions and extract feature from them.

\noindent\textbf{Test Time Adaptation} (Figure \ref{fig:ourapproach}, right). Given the single target image $x^t$, we adapt on it the original backbone's parameters $\theta_{f}$ by finetuning the rotation recognition through
\vspace{-1mm}\begin{equation}
    \argmin_{\theta_{f}, \theta_{r}}  \mathcal{L}_{r}(G_{r}(G_{f}(R(x^t)|\theta_{f})|\theta_{r}),q^t)~.
    \label{eq:finetuning}\vspace{-1mm}
\end{equation}
This process involves only $G_{f}$ and $G_{r}$, while the RPN and ROI detection components described by $G_{d}$ remain unchanged. In the following we use $\gamma$ to indicate the number of gradient steps (\ie iterations), with $\gamma=0$ corresponding to the  pretraining phase. At the end of the finetuning process, the inner feature model is  described by $\theta^*_f$ and the detection prediction on $x^t$ is obtained by $y^{t*} = G_{d}(G_{f}(x^t|\theta^*_{f})|\theta_{d})$. 

We use the name FULL-OSHOT to indicate our new approach.
The meta-learning strategy is summarized in Algorithm \ref{alg:meta}, while the adaptation process on a single target sample is outlined in Algorithm \ref{alg:adapt}.
We also consider two intermediate cases: \emph{Tran}-OSHOT extends OSHOT with the data semantic-preserving transformations used in FULL-OSHOT, and \emph{Meta}-OSHOT corresponds to FULL-OSHOT without transformations (\ie $K=1$).

\begin{figure}[t]
\removelatexerror
\SetNlSty{small}{}{}
\begin{algorithm}[H]
\small
\SetAlgoLined
\KwInput{$G_{f}$, $G_{d}$, $G_{r}$, parameters $\theta_{f}$, $\theta_{d}$,  $\theta_{r}$, rotator $R$, augmenter $A$}
\KwData{Source image $x$ with $y=(b,c)$}
\While{still $k$ augmentations}{
    $x_k \leftarrow A(x)$\\
    $(\theta'_{f}, \theta'_{r}) \leftarrow(\theta_{f}, \theta_{r}) $ ~~\Comment{\texttt{copy params}}\\ 
 \While{still $\eta$ iterations}{
 $\tilde{b},\tilde{c}$ $\leftarrow$ $G_{d}(G_{f}(x_k|\theta'_{f})|\theta_{d})$\\
    $x_{r,k}$ $\leftarrow$ $R(x_k)$ ~~\Comment{\texttt{rand. rotation $q$}}\\ 
    $b_{r,k}$ $\leftarrow$ $R(\tilde{b})$ \\ 
 minimize self-supervised loss\\
 \scriptsize{$(\theta'_{f}, \theta'_{r}) \leftarrow  (\theta'_{f}, \theta'_{r}) - \alpha \nabla_{\theta'_{f}, \theta'_{r}} \mathcal{L}_{r}(G_{r}(G_{f}(b_{r,k}|\theta'_{f})|\theta'_{r}),q) $}  \\\nonl~~\Comment{ $G_{r}(\cdot|\theta'_{r}) = FC_{\theta'_{r}}(\textit{pseudoboxcrop}(\cdot))$}
 }
 compute the supervised loss
 $l_k =\mathcal{L}_{d}(G_{d}(G_{f}(x_{k}|\theta'_{f})|\theta_{d}), y)$
 }
 minimize the supervised loss
 {$(\theta^*_{f}, \theta^*_{d}) \leftarrow  (\theta_{f}, \theta_{d}) - \beta \nabla_{\theta_{f}, \theta_{d}} \sum_{k \in K} l_k$}
 \caption{\small\textbf{Meta-Learning on one source sample}}
 \label{alg:meta}
\end{algorithm}\vspace{-6mm}
\end{figure}
%
\begin{figure}[h!]
\removelatexerror
\SetNlSty{small}{}{}
\begin{algorithm}[H]
\small
\SetAlgoLined
\KwInput{$G_{f}$, $G_{d}$, $G_{r}$, parameters $\theta_{f}$, $\theta_{d}$,  $\theta_{r}$, from the pretraining phase, rotator $R$}
\KwData{Target image $x^t$}
    $(\theta^*_f, \theta^*_r) \gets (\theta_f, \theta_r)$ ~~\Comment{\texttt{copy params}}\\ 
 \While{still $\gamma$ iterations}{
 $\tilde{b}^t,\tilde{c}^t$ $\leftarrow$ $G_{d}(G_{f}(x^t|\theta^*_{f})|\theta_{d})$\\
    $x_r^t$ $\leftarrow$ $R(x^t)$ ~~\Comment{\texttt{rand.  rotation $q$}}\\ 
    $b_r^t$ $\leftarrow$ $R(\tilde{b}^t)$ \\
 minimize self-supervised loss\\
 \scriptsize{$(\theta^*_{f}, \theta^*_{r}) \leftarrow  (\theta^*_{f}, \theta^*_{r}) - \alpha \nabla_{\theta^*_{f}, \theta^*_{r}} \mathcal{L}_{r}(G_{r}(G_{f}(b_r^t|\theta^*_{f})|\theta^*_{r}),q) $} \\\nonl~~\Comment{ $G_{r}(\cdot|\theta^*_{r}) = FC_{\theta^*_{r}}(\textit{pseudoboxcrop}(\cdot))$}
}
final detection prediction using update parameters\\
$y^{t*} = G_{d}(G_{f}(x^t|\theta^*_{f})|\theta_{d})$.
 \caption{\small\textbf{Adaptive phase on one target sample}}
 \label{alg:adapt}
\end{algorithm}\vspace{-5mm}
\end{figure}

\section{Experiments}
\noindent\textbf{Dataset and Competitors.} 
We run an extensive experimental analysis on several datasets.  
The Pascal \emph{Visual Object Classes (VOC)} Pascal-VOC \citep{everingham2010pascal} is a real-world image collection covering bounding boxes annotations for 20 common categories.
The \emph{Artistic Media Datasets (AMD)} is composed of Clipart1k, Comic2k and Watercolor2k \citep{inoue2018cross}. The first shares its 20 categories with VOC. The other two contain a 6 class subset of VOC.
\emph{Cityscapes} \citep{cordts2016cityscapes} is an urban street scene dataset with pixel level annotations of 8 categories from which it is possible to obtain the corresponding bounding boxes as in \citep{Chen_2018_CVPR}.
\emph{Foggy Cityscapes} \citep{sakaridis2018semantic} contains different levels of synthetic fog over Cityscapes. We consider images with the highest amount of artificial fog.

\rev{Finally, \emph{Social Bikes} is our new dataset containing 530 images of scenes with persons/bicycles collected from {Twitter}, {Instagram} and {Facebook} by searching for \emph{\#bike} tags (see the top part of Figure \ref{fig:newteaser}). We designed it to be used as target when the source domain is VOC, indeed the two classes person and bicycles are shared among them. With respect to the other testbeds, Social Bikes covers a larger variety of visual styles related to the tastes and preferences of each social media user. This is quantitatively confirmed by the average standard deviation of the Domain2Vec style features components \citep{domain2vec}: the value for Social Bikes ($2.10 \times 10^{-4}$) is higher with respect to those of the AMDs which share the same VOC source (Clipart $1.99 \times 10^{-4}$, Comic $1.96 \times 10^{-4}$ and Watercolor $1.71 \times 10^{-4}$)}.

Our \emph{Baseline} is Faster-RCNN trained on the source domain and deployed on the target without further adaptation. \emph{\textit{Tran}-Baseline} is a variant obtained by applying at training time the same data semantic-preserving transformations introduced in FULL-OSHOT.
This allows us to assess how much of the improvement is due to data augentation rather than to the training strategy. 
We chose as benchmark methods \emph{DivMatch} \citep{diversifymatch_Kim_2019_CVPR}, \emph{SW} \citep{Saito_2019_CVPR}, \emph{SW-ICR-CCR} \citep{xuCVPR2020} and \emph{ICCR-VDD} \citep{wu2021vector} already described in Section \ref{sec:related}.
In all the cases we use a ResNet-50 backbone pretrained on ImageNet for fair comparison. 

\noindent \rev{\textbf{Implementation and Setting Details.}} 
To run all the experiments we resized the image's shorter side to 600 pixels and apply random horizontal flipping during pretraining. \rev{The weight $\lambda$ is set to $0.05$. Our model is robust to the exact value of this parameter in [0.01, 0.2]: 
the relevance of the rotation recognition objective should be high enough for the auxiliary task to be learned, but low enough to not hijack the main task learning.} 

\rev{The multi-task pretraining stage of OSHOT runs for 
 70k iterations using SGD with momentum set at 0.9, the initial learning rate is 0.001 and decays by a factor 10 after 50k iterations. We use a batch size of 1, keep batch normalization layers fixed for both pretraining and adaptation phases and freeze the first 2 blocks of ResNet50.
 FULL-OSHOT is actually trained in two steps. For the first 60k iterations the training is identical to that of OSHOT, while in the last 10k iterations the meta-learning procedure is activated. The inner loop optimization on the self-supervised task runs with $\eta=5$ iterations and the batch size is 2 to accommodate for two transformations of the original image. Specifically we used gray-scale and color-jitter with brightness, contrast, saturation and hue all set to 0.4. All the other hyperparameters remain unchanged as in OSHOT.
 \emph{Tran}-OSHOT differs from OSHOT only for the last 10k learning iterations, where the batch size is 2 and the network sees images augmented using the same transformations of FULL-OSHOT.
 \emph{Meta}-OSHOT is instead identical to FULL-OSHOT, made exception for the transformations which are dropped, thus the batch size is 1 also in the last 10k pretraining iterations.
}

\begin{figure}[t]
\begin{tabular}{c}
\hspace{-3mm}\includegraphics[width=0.49\textwidth]{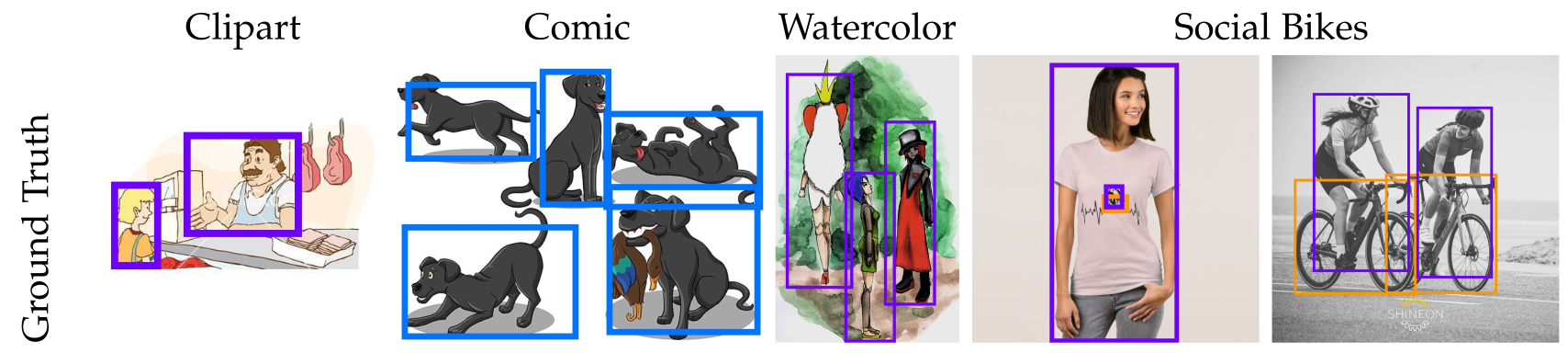}\\
\hspace{-3mm}\includegraphics[width=0.49\textwidth]{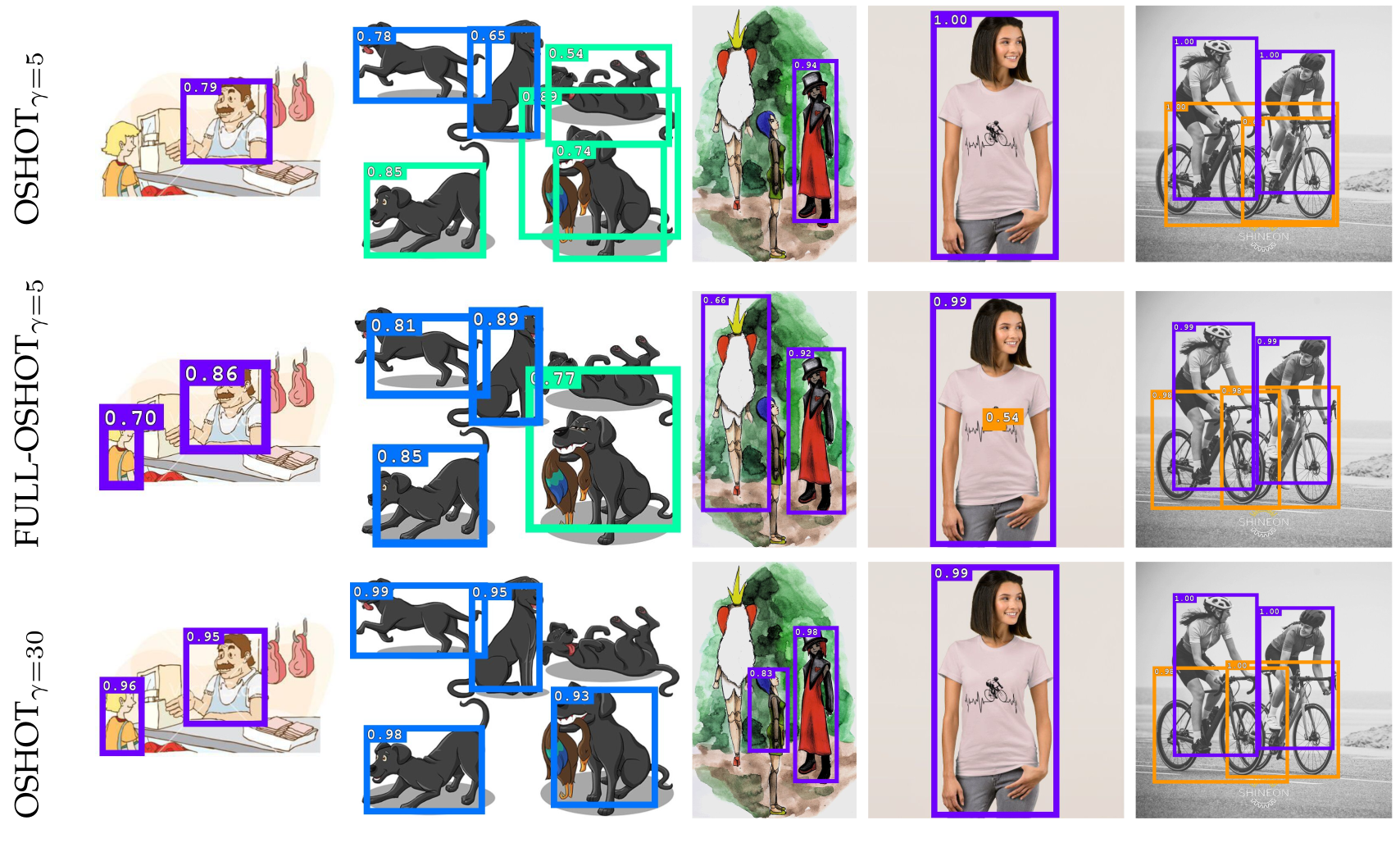}
\end{tabular}\vspace{-3mm}
    \caption{\rev{Some detection results examples of FULL-OSHOT and OSHOT when changing the number of adaptive iterations.}} 
    \label{fig:qualitative} \vspace{-3mm}
\end{figure}

The detection performance (mAP) is assessed with IoU threshold at 0.5. 
In the following we use $Source\rightarrow Target$ to indicate the experimental setting and report the average of three independent runs. 
Our detailed error analysis is obtained via TIDE \citep{tide-eccv2020}:
it estimates how much each type of detection failure contributes to the missing mAP. \rev{Thus, its quantitative insight go beyond standard qualitative visualizations (see some of them in Figures \ref{fig:newteaser}, \ref{fig:qualitative} and in the github page)}.
It counts false positives and false negatives, 
and identifies six error categories. 
\emph{Cls} means object localized correctly ($IoU_{max} \geq 0.5$) but classified incorrectly, \emph{Loc} means object classified correctly but localized incorrectly ($0.1 \leq IoU_{max} < 0.5$). \emph{Both} is used when the two situations occur simultaneously. In \emph{Dupe} the detection is correct, but the same ground truth bounding box was already associated with another higher scoring detection.  \emph{Bkg} means detected background as foreground ($IoU_{max} < 0.1$) and \emph{Miss} is for all the undetected ground truth boxes not covered by other types of errors. 

\rev{
Finally, we remark that none of the state of the art cross-domain detection algorithms used as reference were designed to manage adaptation on a single unlabeled target image, and fail in that condition. We still include those methods in our benchmark by 
favouring them with access during training to ten target images randomly selected at each run (\emph{Ten-Shot Target}), or even to the entire target set (\emph{Whole Target}). 
We collect average precision statistics during inference.
}

\noindent\textbf{Adapting to social feeds.} When the data comes from multiple providers, the  assumption that all target images originate from the same underlying distribution does not hold and standard cross-domain detection methods are penalized regardless of the number of seen target samples. We pretrain the source detector on VOC, and deploy it on Social Bikes.

In Table \ref{table:social} the mAP results with $\gamma=0$ allow us to 
compare the pretraining models before adaptation and already show the advantage of FULL-OSHOT over OSHOT, as well as over the \emph{Tran} and \emph{Meta} variants. When $\gamma=5$ all variants of OSHOT obtain an improvement that ranges from $1.9$ (OSHOT) to $2.6$ (FULL-OSHOT) points over the Baseline just by adapting on a single test sample. 
Despite granting them access to the whole set of adaptation samples, the reference domain adaptive algorithms reach at best an advantage of $1.2$ points over FULL-OSHOT. 
When using ten target samples, half of the methods show a negative transfer with respect to 
the Baseline.  

By looking at the detection error analysis we can see that the adaptation iterations allow OSHOT to reduce the number of false negatives. Moreover, both \textit{Tran}-OSHOT and FULL-OSHOT obtain a higher mAP than OSHOT thanks to lower \textit{Miss} errors. 
The performance of FULL-OSHOT confirms that the meta-learning strategy with semantic-preserving data augmentations successfully prepares the model to solve the adaptation task at inference time.

\begin{table}[tb]
\small
\centering
\begin{tabular}{clcc|c} 
\hline
\multicolumn{5}{c}{\textsl{\textbf{One-Shot} Target}}\\
\hline
\multicolumn{2}{c}{Method} & person & bicycle & mAP\\ \hline
\multicolumn{2}{c}{Baseline} &  69.0 & 74.1 & 71.6 \\ 
\multicolumn{2}{c}{\emph{Tran}-Baseline} &  71.4 & 74.2 & 72.8 \\ 
\hline
\multirow{4}{*}{$\gamma = 0$} & {OSHOT}  & 68.9 & 74.6 & 71.8 \\
& \emph{Tran}{-OSHOT}  & 71.6 & 74.0 &  72.8\\

& \emph{Meta}{-OSHOT}  & 69.5 & 73.5 &  71.5\\

& {FULL-OSHOT}  & 71.7 & 74.3 &  73.0\\
\hline
\multirow{4}{*}{$\gamma = 5$} & {OSHOT} & 72.1 & 74.9 &  73.5\\
& \emph{Tran}{-OSHOT}  & 73.0 & 74.7 &  73.9\\
& \emph{Meta}{-OSHOT} & 72.6 &  74.5 &  73.6\\
& {FULL-OSHOT}  &  73.3 & 75.1 &  74.2\\
\hline\hline
\multicolumn{5}{c}{\textsl{\textbf{Ten-Shot} Target}}\\\hline 
\multicolumn{2}{c}{DivMatch  \citep{diversifymatch_Kim_2019_CVPR}} & 69.5 & 73.1 & 71.3 \\
\multicolumn{2}{c}{SW \citep{Saito_2019_CVPR}}  & 69.4 & 73.0  & 71.2  \\
\multicolumn{2}{c}{SW-ICR-CCR \citep{xuCVPR2020}}  & 72.5  & 77.6  & 75.1  \\ 
\multicolumn{2}{c}{VDD-DAOD \citep{wu2021vector}}  & 68.8  & 75.3  & 72.1  \\ 
\hline
\multicolumn{5}{c}{\textsl{\textbf{Whole} Target}}\\
\hline
\multicolumn{2}{c}{DivMatch  \citep{diversifymatch_Kim_2019_CVPR}} & 73.6 & 77.1 &  75.4\\
\multicolumn{2}{c}{SW \citep{Saito_2019_CVPR}}  & 68.6 & 70.3  & 69.5  \\
\multicolumn{2}{c}{SW-ICR-CCR \citep{xuCVPR2020}}  &  72.0 &  72.8 & 72.4  \\ 
\multicolumn{2}{c}{ICCR-VDD \citep{wu2021vector}}  &  71.1 &  71.9 & 71.5  \\ 
\hline
\end{tabular}
\begin{tabular}{@{}c@{~~}c@{~~}c}
\hspace{5mm}\small{Baseline} & \small{OSHOT$_{\gamma=0}$} & \small{OSHOT$_{\gamma=5}$}\\
\hspace{5mm}\includegraphics[width=0.14\textwidth]{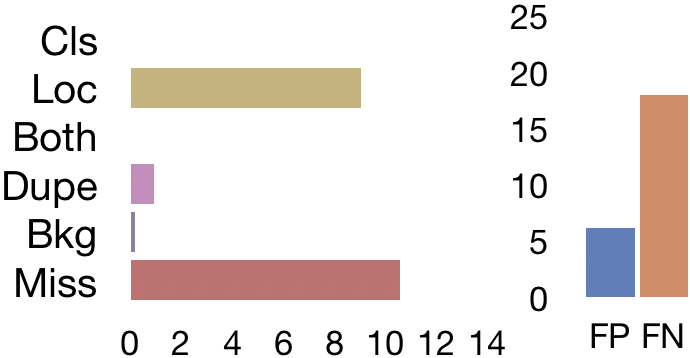} & 
\includegraphics[width=0.14\textwidth]{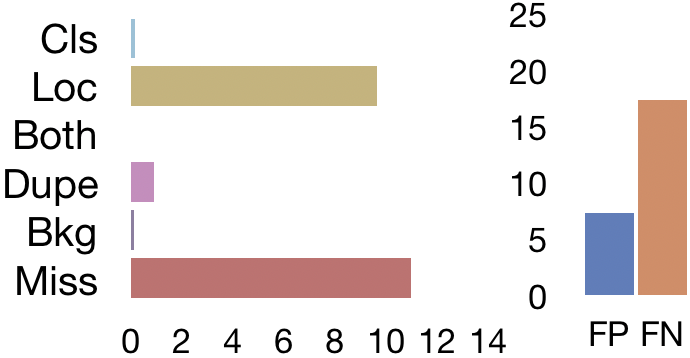} &
\includegraphics[width=0.14\textwidth]{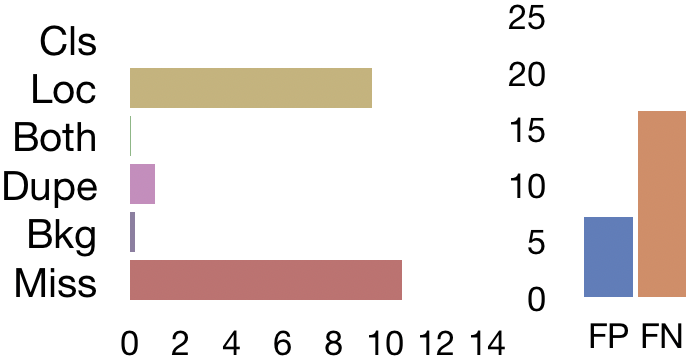} \\
\hspace{5mm}\small{\emph{Tran}-OSHOT$_{\gamma=5}$} & \small{\emph{Meta}-OSHOT$_{\gamma=5}$} & \small{FULL-OSHOT$_{\gamma=5}$}\\
\hspace{5mm}\includegraphics[width=0.14\textwidth]{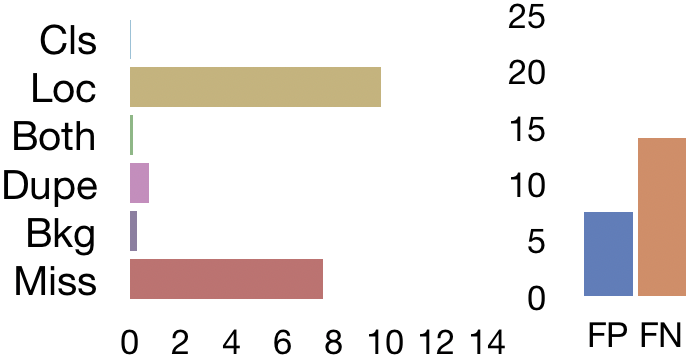} & 
\includegraphics[width=0.14\textwidth]{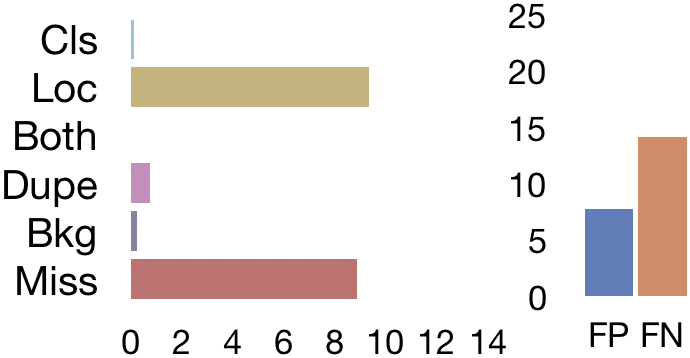} &
\includegraphics[width=0.14\textwidth]{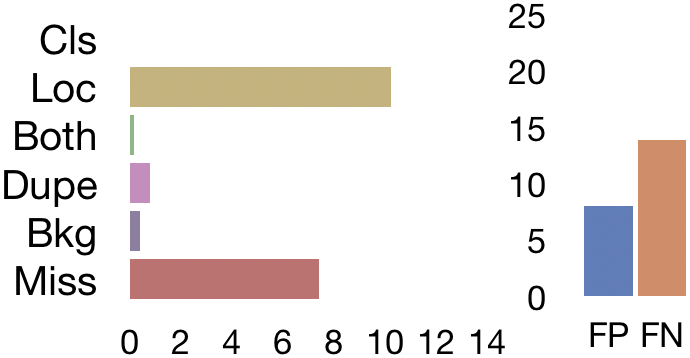} \\
\end{tabular}
\caption{Results for VOC $\rightarrow$ Social Bikes. The histograms illustrate the detection error analysis performed with TIDE \citep{tide-eccv2020}.}
\label{table:social}\vspace{-5mm}
\end{table}
\begin{table}[h!]
\small
\centering
\begin{adjustbox}{width=0.5\textwidth}
\subfloat[VOC $\rightarrow$ Clipart]{
{}
\begin{tabular}{c@{~~}l@{~~}c@{~~}c@{~~}}
\hline
\multicolumn{3}{c}{\textsl{\textbf{One-Shot} Target}}\\
\hline
\multicolumn{2}{c}{Method} & mAP\\ \hline
\multicolumn{2}{c}{Baseline} & 26.4 \\
\multicolumn{2}{c}{\emph{Tran}-Baseline} &  27.6 \\ \hline
\multirow{4}{*}{$\gamma = 0$} & OSHOT  & 28.8 \\
& \emph{Tran}-OSHOT  &  28.6\\
& \emph{Meta}-OSHOT  &  29.4\\
& FULL-OSHOT  &  28.6\\
\hline
\multirow{4}{*}{$\gamma = 5$} & OSHOT  &  30.8 \\
& \emph{Tran}-OSHOT &  30.5\\
& \emph{Meta}-OSHOT  &  31.4\\
& FULL-OSHOT  &  31.7\\
\hline\hline
\multicolumn{3}{c}{\textsl{\textbf{Ten-Shot} Target}}\\
\hline
\multicolumn{2}{c}{DivMatch  \citep{diversifymatch_Kim_2019_CVPR}} &  26.3 \\ 
\multicolumn{2}{c}{SW \citep{Saito_2019_CVPR}} &  26.4 \\ 
\multicolumn{2}{c}{SW-ICR-CCR \citep{xuCVPR2020}} &  27.2 \\
\multicolumn{2}{c}{ICCR-VDD \citep{wu2021vector}} &  27.6 \\\hline
\end{tabular}
}
\subfloat[VOC $\rightarrow$ Comic]{
\begin{tabular}{c@{~~}}
\hline
\multicolumn{1}{c}{\textsl{\textbf{One-Shot} Target}}\\
\hline
mAP\\ \hline
18.1 \\
22.4  \\ \hline
19.9 \\
20.1\\
20.2\\
21.1\\ \hline
22.3 \\
24.9\\
24.8\\
25.2\\
\hline\hline
\multicolumn{1}{c}{\textsl{\textbf{Ten-Shot} Target}}\\
\hline
20.8 \\ 
21.0 \\
21.1 \\
24.8 \\
\hline
\end{tabular}
} 
\subfloat[VOC $\rightarrow$ Watercolor]{
\begin{tabular}{c@{~~}}
\hline
\multicolumn{1}{c}{\textsl{\textbf{One-Shot} Target}}\\
\hline
mAP\\ \hline
42.8 \\
46.3 \\\hline
45.7\\
45.4\\
45.8\\
46.4\\
\hline
48.1 \\
47.7\\
49.0\\
48.9 \\
\hline\hline
\multicolumn{1}{c}{\textsl{\textbf{Ten-Shot} Target}}\\
\hline
45.4 \\
42.0 \\
45.3 \\
43.1 \\
\hline
\end{tabular}
}
\end{adjustbox}
\begin{adjustbox}{width=0.5\textwidth}
\begin{tabular}{p{0.02\textwidth}@{~~}p{0.17\textwidth}@{~~}p{0.17\textwidth}@{~~}p{0.17\textwidth}}
\multirow{4}{*}{(a)} & \hspace{4mm} Baseline & \hspace{3mm} OSHOT$_{\gamma=0}$ & \hspace{3mm} OSHOT$_{\gamma=5}$ \\
 & \raisebox{-.5\height}{\includegraphics[width=0.17\textwidth]{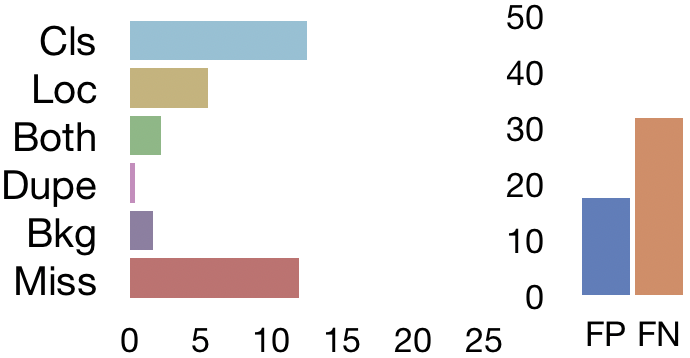}} & \raisebox{-.5\height}{\includegraphics[width=0.17\textwidth]{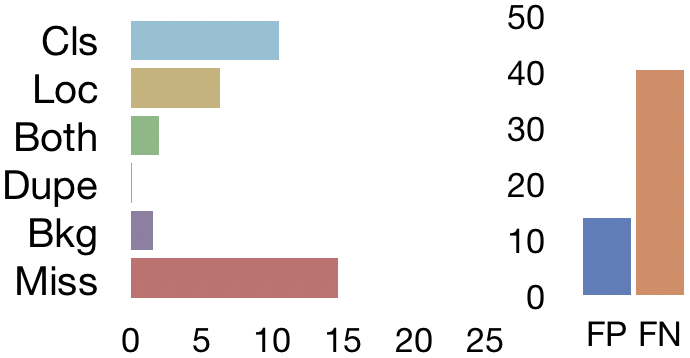}} &
\raisebox{-.5\height}{\includegraphics[width=0.17\textwidth]{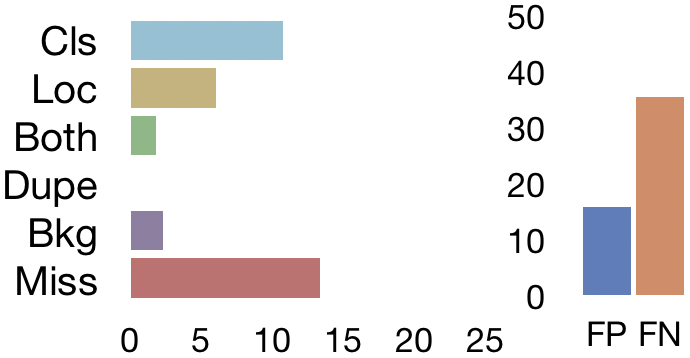}} \\
 & \emph{Tran}-OSHOT$_{\gamma=5}$ & \emph{Meta}-OSHOT$_{\gamma=5}$ & FULL-OSHOT$_{\gamma=5}$\\
& \raisebox{-.5\height}{\includegraphics[width=0.17\textwidth]{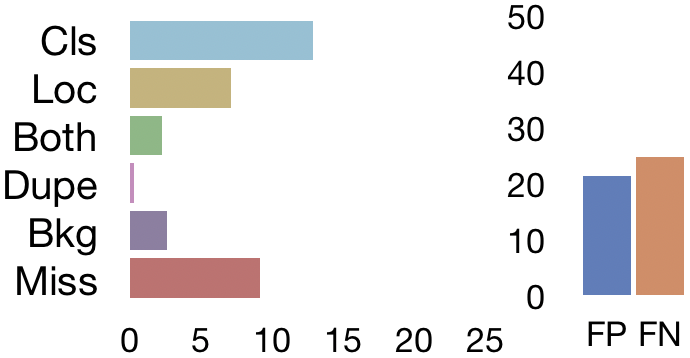}} & 
\raisebox{-.5\height}{\includegraphics[width=0.17\textwidth]{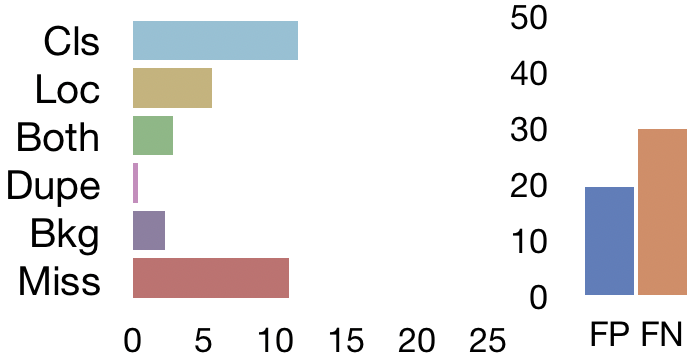}} &
\raisebox{-.5\height}{\includegraphics[width=0.17\textwidth]{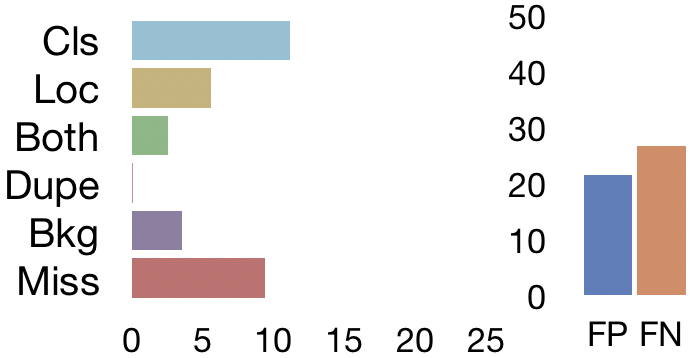}} \vspace{1mm}\\
\multirow{4}{*}{(b)}  & \hspace{4mm} Baseline & \hspace{3mm} OSHOT$_{\gamma=0}$ & \hspace{3mm} OSHOT$_{\gamma=5}$ \\
& \raisebox{-.5\height}{\includegraphics[width=0.17\textwidth]{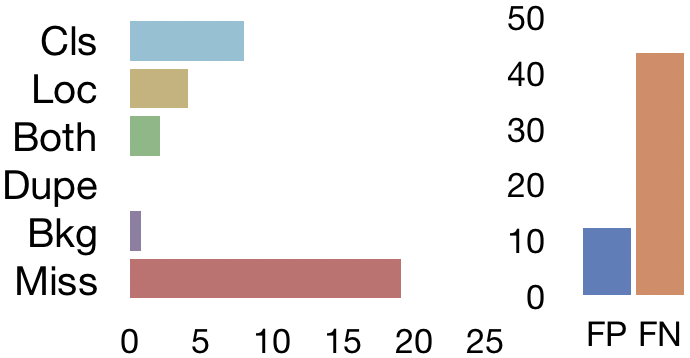}} & 
\raisebox{-.5\height}{\includegraphics[width=0.17\textwidth]{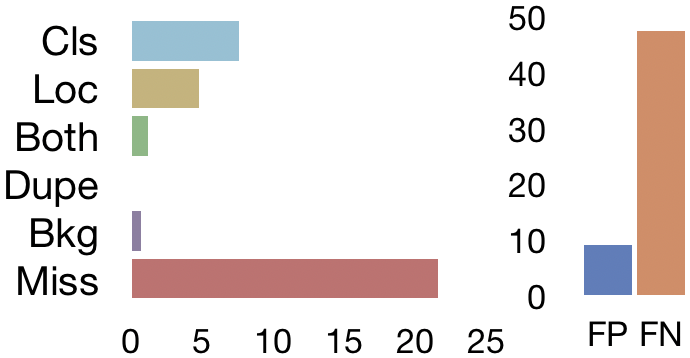}} &
\raisebox{-.5\height}{\includegraphics[width=0.17\textwidth]{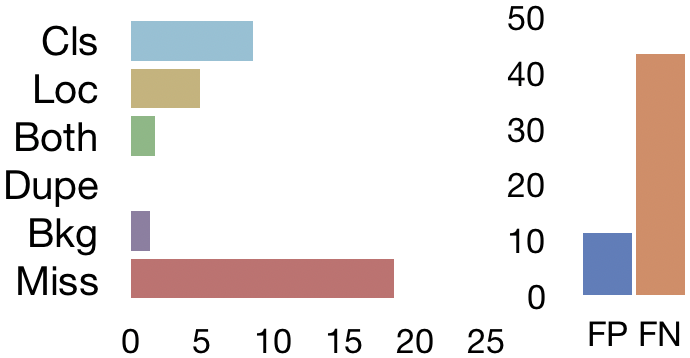}} \\
& \emph{Tran}-OSHOT$_{\gamma=5}$ & \emph{Meta}-OSHOT$_{\gamma=5}$ & FULL-OSHOT$_{\gamma=5}$\\
&\raisebox{-.5\height}{\includegraphics[width=0.17\textwidth]{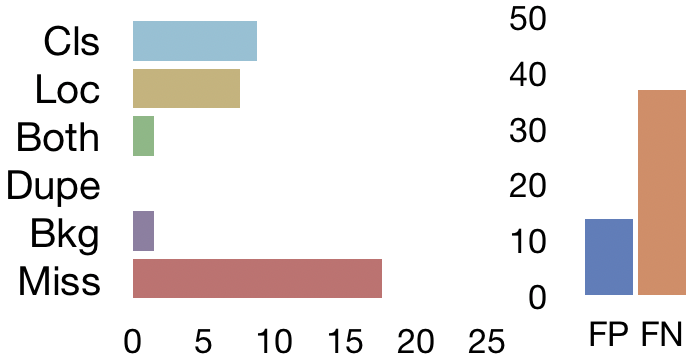}} & 
\raisebox{-.5\height}{\includegraphics[width=0.17\textwidth]{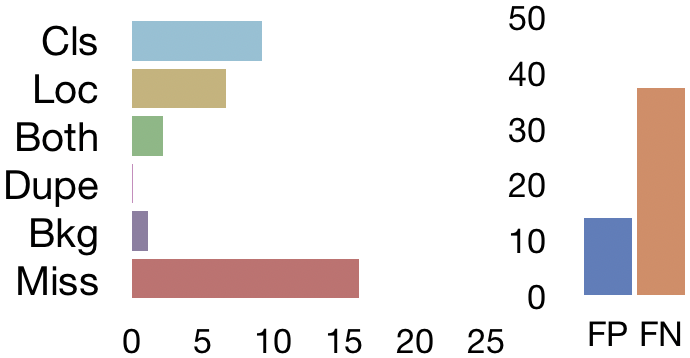}} &
\raisebox{-.5\height}{\includegraphics[width=0.17\textwidth]{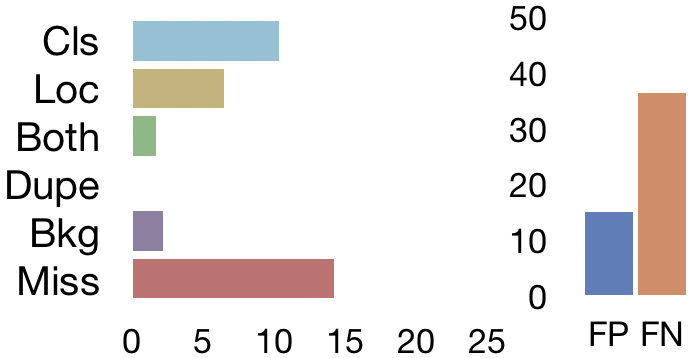}} \\
\multirow{4}{*}{(c)}  & \hspace{4mm} Baseline & \hspace{3mm} OSHOT$_{\gamma=0}$ & \hspace{3mm} OSHOT$_{\gamma=5}$ \\
& \raisebox{-.5\height}{\includegraphics[width=0.17\textwidth]{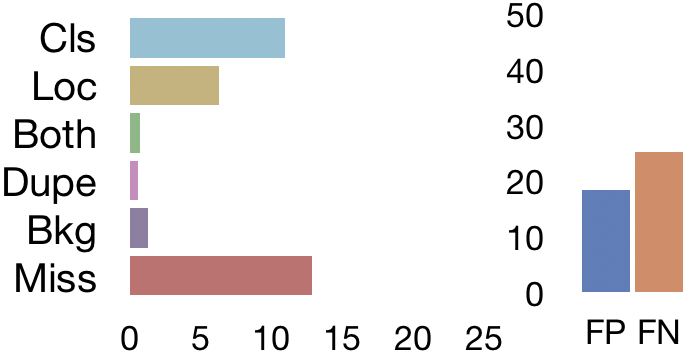}} & 
\raisebox{-.5\height}{\includegraphics[width=0.17\textwidth]{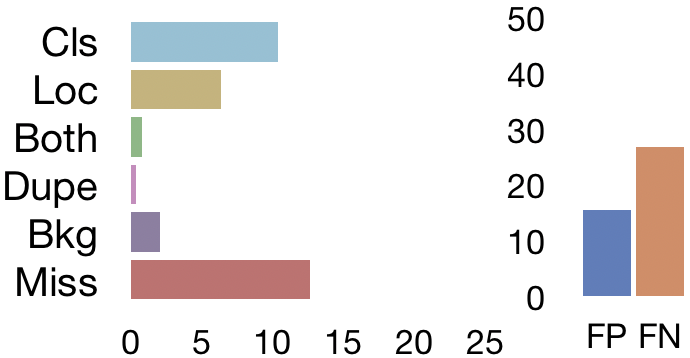}} &
\raisebox{-.5\height}{\includegraphics[width=0.17\textwidth]{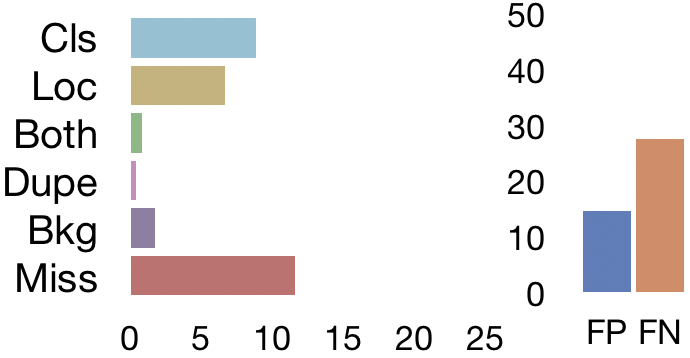}} \\
& \emph{Tran}-OSHOT$_{\gamma=5}$ & \emph{Meta}-OSHOT$_{\gamma=5}$ & FULL-OSHOT$_{\gamma=5}$\\
&\raisebox{-.5\height}{\includegraphics[width=0.17\textwidth]{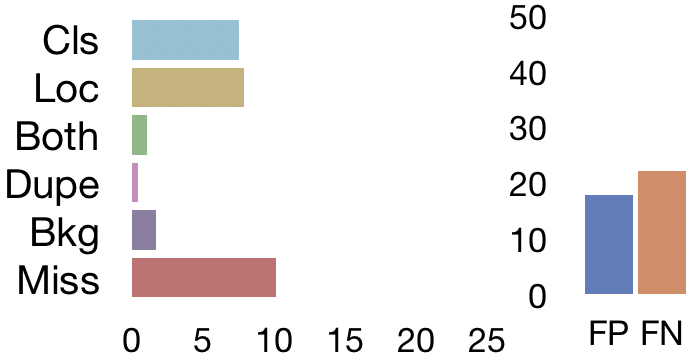}} & 
\raisebox{-.5\height}{\includegraphics[width=0.17\textwidth]{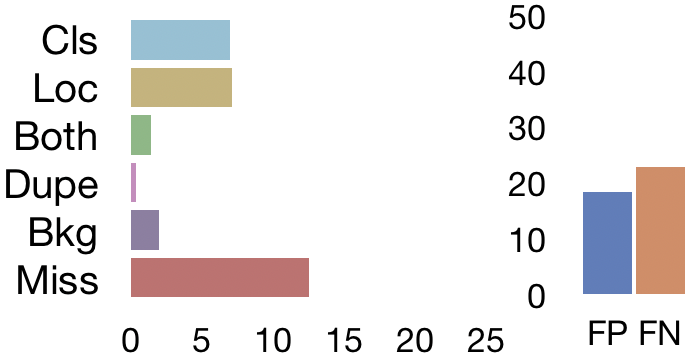}} &
\raisebox{-.5\height}{\includegraphics[width=0.17\textwidth]{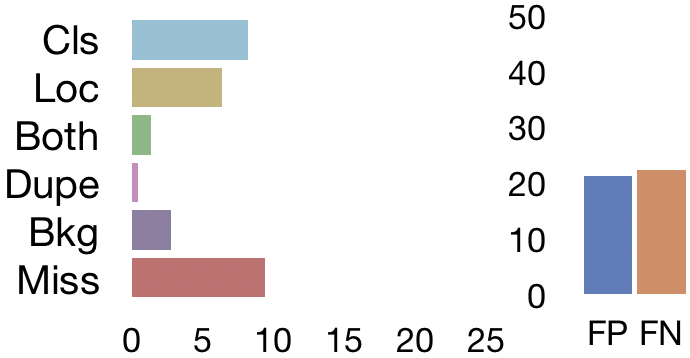}} \\
\end{tabular}
\end{adjustbox}
\caption{Results for VOC $\rightarrow$ AMD and corresponding histograms with the error analysis performed with TIDE \citep{tide-eccv2020}.}
\label{table:VOC2AMD}\vspace{-3mm}
\end{table}

\begin{table}[tb]
\centering
\begin{adjustbox}{width=0.48\textwidth}
\centering
\begin{tabular}{c@{~~}l@{~~}c@{~~}c@{~~}c@{~~}c@{~~}c@{~~}c@{~~}c@{~~}c@{~~}|c@{~~}}
\hline
\multicolumn{11}{c}{\textsl{\textbf{One-Shot} Target}}\\
\hline
\multicolumn{2}{c}{Method} & person & rider & car & truck & bus & train & mcycle & bicycle & mAP\\ \hline
\multicolumn{2}{c}{Baseline} & 30.4 & 36.3 & 41.4 & 18.5  & 32.8 & 9.1 & 20.3 & 25.9 & 26.8 \\
\multicolumn{2}{c}{\emph{Tran}-Baseline} & 32.1 & 35.2 & 42.9 & 17.8 & 31.0 & 4.3 & 22.6 & 30.0 & 27.0 \\ \hline
\multirow{4}{*}{$\gamma = 0$} & OSHOT  & 32.2 & 38.6 & 39.0 & 20.5 & 30.6 & 12.9 & 22.4 & 31.2 & 28.4 \\
& \emph{Tran}-OSHOT  & 30.5 & 37.4 & 42.7 & 16.9 & 29.5 & 14.5 & 21.9 & 30.4 & 28.0\\
& \emph{Meta}-OSHOT & 30.6 & 35.1 & 35.9 & 16.6 & 28.4 & 7.6 & 18.2 & 28.4 & 25.1\\
& FULL-OSHOT  & 31.7 & 40.8 & 43.7 & 18.3 & 28.8 & 11.0 & 22.8 & 33.3 & 28.8\\
\hline
\multirow{4}{*}{$\gamma = 5$} & OSHOT  & 32.7 & 39.3 & 41.1 & 21.1 & 33.1 & 12.6 & 22.7 & 31.9 & 29.3 \\
& \emph{Tran}-OSHOT  & 30.9 & 38.5 & 43.0 & 17.5 & 32.1 & 13.9 & 21.6 & 30.5 & 28.5\\
& \emph{Meta}-OSHOT  & 32.1 & 38.2 & 39.9 & 17.4 & 30.9 & 7.5 & 21.0 & 29.2 & 27.0\\
& FULL-OSHOT  & 32.0 & 39.7 & 43.8 & 18.8 & 31.8 & 10.6 & 22.1 & 33.2 & 29.0\\
\hline\hline
\multicolumn{11}{c}{\textsl{\textbf{Ten-Shot} Target}}\\
\hline
\multicolumn{2}{c}{DivMatch  \citep{diversifymatch_Kim_2019_CVPR}} & 27.6 & 38.1 & 42.9 & 17.1 & 27.6 & 14.3 & 14.6 & 32.8 & 26.9 \\
\multicolumn{2}{c}{SW \citep{Saito_2019_CVPR}}  & 25.5 & 30.8 & 40.4 & 21.1 & 26.1 & 34.5 & 6.1 & 13.4 & 24.7 \\
\multicolumn{2}{c}{SW-ICR-CCR \citep{xuCVPR2020}} & 29.6  & 40.8 & 39.6 & 20.5 & 32.8 & 11.1 &  24.0 & 34.0 & 29.1 \\
\multicolumn{2}{c}{ICCR-VDD \citep{wu2021vector}} & 32.3  & 32.1 & 41.7 & 25.0 & 29.0 & 40.0 &  12.6 & 19.7 & 29.0 \\
\hline
\end{tabular}
\end{adjustbox}\hspace{5mm}
\begin{tabular}{@{}c@{~~}c@{~~}c}
\hspace{5mm}\small{Baseline} & \small{OSHOT$_{\gamma=0}$} & \small{OSHOT$_{\gamma=5}$}\\
\hspace{5mm}\includegraphics[width=0.14\textwidth]{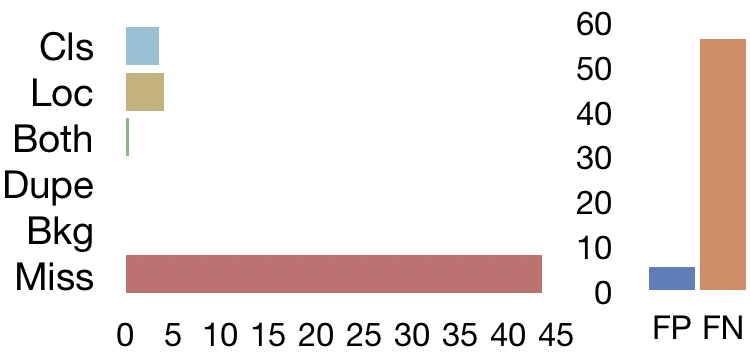} & 
\includegraphics[width=0.14\textwidth]{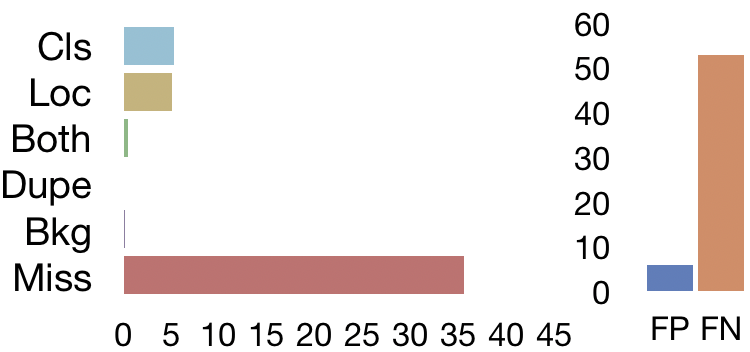} &
\includegraphics[width=0.14\textwidth]{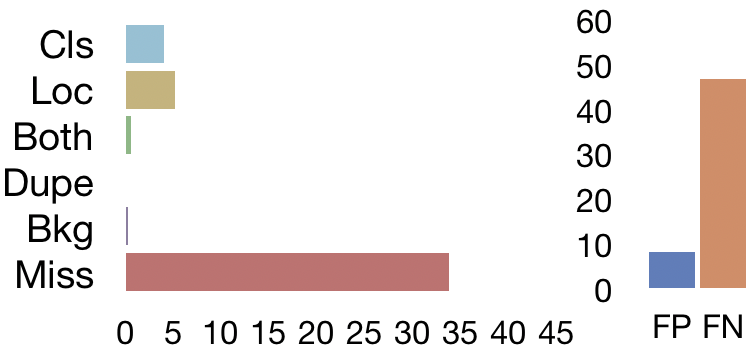} \\
\hspace{5mm}\small{\emph{Tran}-OSHOT$_{\gamma=5}$} & \small{\emph{Meta}-OSHOT$_{\gamma=5}$} & \small{FULL-OSHOT$_{\gamma=5}$}\\
\hspace{5mm}\includegraphics[width=0.14\textwidth]{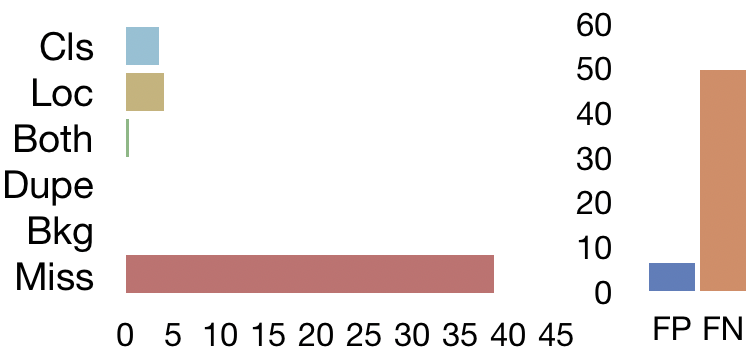} & 
\includegraphics[width=0.14\textwidth]{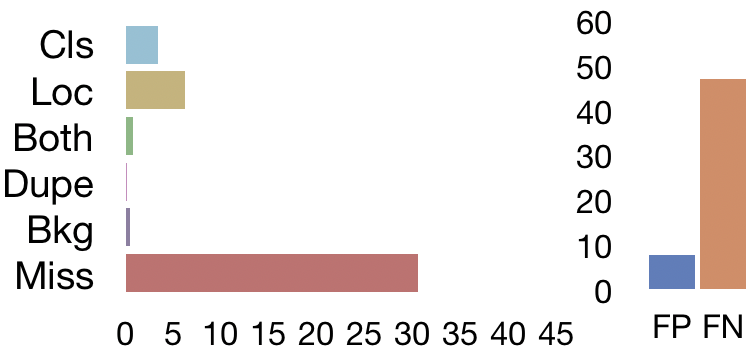} &
\includegraphics[width=0.14\textwidth]{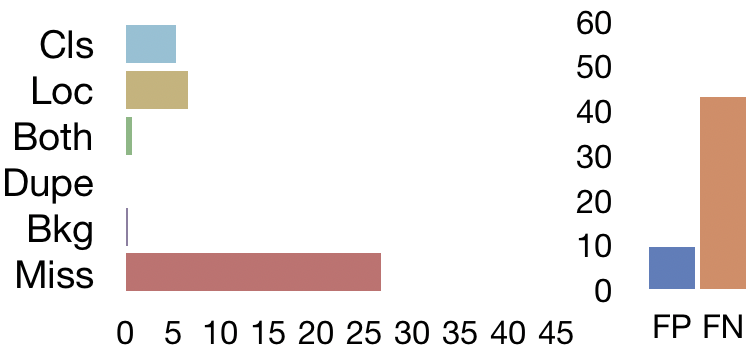} \\
\end{tabular}
\caption{Results for Cityscapes $\rightarrow$ FoggyCityscapes. The histograms illustrate the detection error analysis performed with TIDE \citep{tide-eccv2020}.}
\label{table:C2F}\vspace{-3mm}
\end{table}

\noindent\textbf{Large distribution shifts.} Artistic images are difficult testbed for cross-domain methods. Unpredictable perturbations in shape and color are challenging to detectors trained only on real world photos. We investigate this setting by training on VOC and testing on Clipart, Comic and Watercolor datasets.
\rev{Table \ref{table:VOC2AMD} shows that OSHOT and its variants, by only exploiting one sample at a time with few adaptive iterations ($\gamma=5$), outperform the adaptive detectors which leverage on ten target samples.}
More precisely, none of the adaptive detectors are able to work in data scarcity conditions and obtain results comparable to those of the \emph{Tran}-Baseline and of the pretraining phase of our approach ($\gamma=0$).
We also highlight that when $\gamma=5$, \emph{Meta}-OSHOT obtains results higher than \emph{Tran}-OSHOT and only slightly lower on average than FULL-OSHOT, thus the meta-learning strategy alone (without additional data augmentation) prepares the detector to the inference time adaptation. 

From the detection error analysis we see that the data augmentation of \emph{Tran}-OSHOT pushes for a lower number of errors of type \textit{Miss}, while the meta learning strategy of \emph{Meta}-OSHOT gets a lower number of \textit{Classification} errors. FULL-OSHOT takes advantage of both, obtaining the best performance.

\noindent\textbf{Adverse weather.} 
Some environmental conditions, such as fog, may be disregarded in source data acquisition, yet adaptation to these circumstances is crucial in real world. 
We consider the Cityscapes $\rightarrow$ FoggyCityscapes setting by training our base detector on the first domain.
We perform model selection on the Cityscapes validation split before deployment. 

The results in Table \ref{table:C2F} show that domain adaptive detectors struggle in this scenario. 
Only SW-ICR-CCR and VDD-DAOD are able to exploit the small adaptation set and obtain a meaningful improvement over the Baseline. 
For what concerns OSHOT and its variants, the pretraining alone ($\gamma=0$) helps in gaining a better generalization ability: all variants but \emph{Meta}-OSHOT show higher performance than the Baseline. The advantage is also visible from the \textit{Miss} error type which decreases when passing from the Baseline to OSHOT $\gamma=0$, reaching its lower value for FULL-OSHOT with $\gamma=5$. 

\noindent\textbf{Comparison with One-Shot Style Transfer.}
Although not designed for cross-domain detection, it is possible to apply one-shot style transfer methods as an alternative solution for our setting. We use BiOST \citep{Cohen_2019_ICCV}, to modify the style of the target sample towards that of the source domain before performing inference. Due to the time-heavy requirements to perform BiOST on each test sample, we test it on Social Bikes and on a random subset of 100 Clipart images that we name Clipart100. We compare performance and time requirements of our approach and BiOST on these two targets. 

\begin{table}[tb]
\centering
\begin{adjustbox}{width=0.48\textwidth}
\begin{tabular}{c@{~~}c@{~~}c@{~~}c@{~~}c@{~~}}
\hline
 & \multirow{2}{*}{Baseline} & BiOST & OSHOT & FULL-OSHOT\\ 
 &  & \citep{Cohen_2019_ICCV} & $\gamma=5$ & $\gamma=5$\\ \hline
mAP on Clipart100 & 27.9 & 29.8 & 28.2 & 30.4\\ \hline
mAP on Social Bikes & 71.6 & 71.4 &  73.5  & 74.2\\ \hline
Adaptation time (s per sample) & \multirow{1}{*}{-} & \multirow{1}{*}{$2.4\times10^{4}$}  & \multirow{1}{*}{1.3} & \multirow{1}{*}{1.3}\\
\hline
\end{tabular}
\end{adjustbox}
\caption{Comparison between baseline, one-shot style transfer and our approach in the one-shot unsupervised cross-domain detection setting. Speed computed on an RTX2080Ti with full precision settings.}
\label{table:biost}\vspace{-3mm}
\end{table}

\begin{figure}[t]
    \centering
    \includegraphics[width=0.49\textwidth]{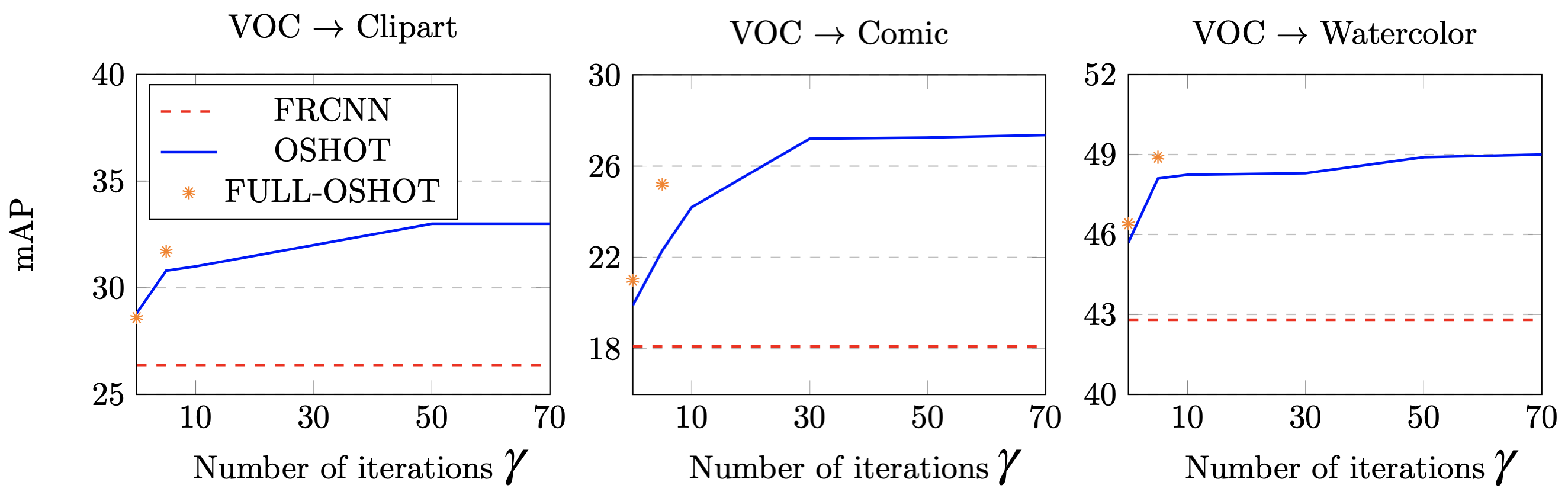}
    \caption{Performance of OSHOT at different number of adaptive iterations.} 
    \label{fig:iterations} \vspace{-3mm}
\end{figure}

Table \ref{table:biost} shows that on Clipart100 the Baseline obtains $27.9$ mAP and BiOST has an advantage over it of $1.9$ points.
Instead on Social Bikes, BiOST incurs in a slight negative transfer, which evidences its inability to effectively modify the source's style on this more challenging testbed. 
OSHOT improves over the baseline on Clipart100 but its mAP remains lower than that of BiOST, while it outperforms both the baseline and BiOST on Social Bikes. 
Finally, FULL-OSHOT shows the best results on both the datasets. The last row of the table presents the time complexity of all the considered methods, which is identical for OSHOT and FULL-OSHOT since the number of adaptive iterations is the same. BiOST instead, needs more than six hours to modify the style of a single source instance. Moreover we highlight that BiOST works under the strict assumption of accessing at the same time the entire source training set and the target sample. Considering these weaknesses and the obtained results, we argue that existing one-shot translation methods are not suitable for one shot unsupervised cross-domain adaptation.

\noindent\textbf{Increasing the number of Adaptive Iterations.}
The bi-level optimization process 
of meta-learning requires non-trivial computational and memory burdens that \rev{might} limit the feasible number of iterations $\eta$. 
In FULL-OSHOT we use the same conditions for the meta pre-training and test time adaptation phases, thus we 
set a small number of training steps with $\gamma=\eta$.
\rev{This however, does not limit the effectiveness of the method, which becomes clear when comparing it with OSHOT at an increasing number of iterations.}
We studied the mAP performance 
on the AMD dataset and collected the results in Figure \ref{fig:iterations}.
We observe a positive correlation between the number of finetuning iterations and the mAP of the model in the earliest steps, while the performance generally reaches a plateau after about 30 iterations: increasing $\gamma$ beyond this value does not affect significantly the final results.
\rev{From the plots we can see that the performance of FULL-OSHOT with just 5 adaptation iterations can be achieved and eventually surpassed by the standard OSHOT only at the cost of a much higher number of adaptation iterations.
This behaviour is also reflected by the visualizations in Figure \ref{fig:qualitative} where the results obtained by FULL-OSHOT with $\gamma=5$ are more similar to those obtained by OSHOT with $\gamma=30$ than those obtained by OSHOT with $\gamma=5$.}

\section{Conclusion}
\noindent This paper focused on {one-shot unsupervised cross-domain detection}, where adaptation should be performed on one single image at inference time, without access to the source data. This scenario holds in  several real world applications like social media monitoring. 
We showed how our FULL-OSHOT outperforms several cross-domain detection methods and improves over its basic OSHOT version \citep{oshot_eccv20} thanks to a novel meta-learning formulation applied on the auxiliary self-supervised task.  This procedure simulates single-sample cross-domain learning episodes and improves the generalization abilities of the detector.

\smallskip\noindent\rev{\textbf{Acknowledgements} Computational resources for this work were provided by IIT (HPC infrastructure).  We would like to thank the anonymous reviewers for their insightful comments.}

\bibliographystyle{model2-names}
\bibliography{ebib}

\end{document}